\documentclass{article}

\usepackage{microtype}
\usepackage{graphicx}
\usepackage{subfigure}
\usepackage{booktabs} 
\usepackage{xcolor}
\definecolor{linkcolor}{RGB}{0,128,255}
\usepackage[colorlinks=true,allcolors=linkcolor,pageanchor=true,plainpages=false,pdfpagelabels,bookmarks,bookmarksnumbered]{hyperref}
\newcommand{\blue}[1]{\textcolor{blue}{#1}}
\newcommand{\airlinesize}{800000}

\usepackage[T1]{fontenc}    
\usepackage{nicefrac}       
\usepackage{amsmath,amssymb,amsfonts}
\usepackage{wrapfig}
\usepackage{booktabs}
\usepackage{mathtools}
\usepackage{dsfont}
\usepackage{commands}
\usepackage{natbib}
\usepackage{setspace}
\usepackage{xspace}
\usepackage[nameinlink]{cleveref}
\usepackage{amsthm}
\usepackage{caption,subfigure}
\usepackage{lscape}

\usepackage{todonotes}

\usepackage{algorithm}
\usepackage{algorithmic}
\newtheorem{theorem}{Theorem}
\newtheorem{proposition}[theorem]{Proposition}

\DeclareMathOperator*{\argmin}{arg\,min}

\usepackage{hyperref}

\usepackage[accepted]{icml2020}

\icmltitlerunning{Healing Products of Gaussan Processes}

\begin{document}

\twocolumn[
\icmltitle{Healing Products of Gaussian Processes}
\icmlsetsymbol{equal}{*}

\begin{icmlauthorlist}
\icmlauthor{Samuel Cohen}{equal,ucl}
\icmlauthor{Rendani Mbuvha}{equal,uj,wits}
\icmlauthor{Tshilidzi Marwala}{uj}
\icmlauthor{Marc Peter Deisenroth}{ucl,uj}
\end{icmlauthorlist}
 
\icmlaffiliation{ucl}{Centre for Artificial Intelligence,  University College London, UK}
\icmlaffiliation{uj}{Institute of Intelligent Systems, University of Johannesburg, South Africa}
\icmlaffiliation{wits}{School of Statistics and Actuarial Science, University of the Witwatersrand, South Africa}

\icmlcorrespondingauthor{Samuel Cohen}{samuel.cohen.19@ucl.ac.uk}
\icmlcorrespondingauthor{Rendani Mbuvha}{rendani.mbuvha@wits.ac.za}
 
%
%
%
 
\vskip 0.3in
]
 

\printAffiliationsAndNotice{\icmlEqualContribution} 

\begin{abstract}
Gaussian processes (GPs) are nonparametric Bayesian models that have been applied to regression and classification problems. One of the approaches to alleviate their cubic training cost is the use of local GP experts trained on subsets of the data. In particular, product-of-expert models combine the predictive distributions of local experts through a tractable product operation. While these expert models allow for massively distributed computation, their predictions typically suffer from erratic behaviour of the mean or uncalibrated uncertainty quantification. By calibrating predictions via a tempered softmax weighting, we provide a solution to these problems for multiple product-of-expert models, including the generalised product of experts and the robust Bayesian committee machine. Furthermore, we leverage the optimal transport literature and propose a new product-of-expert model that combines predictions of local experts by computing their Wasserstein barycenter, which can be applied to both regression and classification.
\end{abstract}

\section{Introduction}
Gaussian processes (GPs) \cite{Rasmussen:2005:GPM:1162254} are nonparametric stochastic processes that have been applied extensively to regression and classification problems. However, their cubic training and quadratic prediction cost hinders their application in large-scale problems. Different approaches alleviate this issue, including sparse approximations~\cite{NIPS2005_2857,soton259182,10.5555/1046920.1194909,Titsias09variationallearning}, the exploitation of structural assumptions \cite{pmlr-v37-wilson15} and local-expert models \cite{10.5555/3008751.3008843,DBLP:conf/nips/RasmussenG01,Cao2014GeneralizedPO,pmlr-v37-deisenroth15,rulliere:hal-01345959,trapp2019structured}. 

Sparse approximations effectively reduce the rank of the covariance matrix through inducing inputs, reducing the training cost from $O(n^{3})$ to $O(nm^{2})$, where $m$ is the number of inducing points and $n$ is the size of the training dataset. Optimisation consists of jointly learning kernel hyperparameters and inducing locations. In particular, \citet{Titsias09variationallearning} treats inducing locations as variational parameters and optimises them and the kernel hyperparameters by maximising a lower bound on the marginal likelihood. \citet{Hensman:2013:GPB:3023638.3023667} scale this approach by introducing mini-batching, reducing the complexity to $O(m^{3})$, while \citet{NIPS2014_5593} reparametrise the problem to allow for distributed inference. 
 
An alternative to sparse GP approximations is to use local experts. Here, the training dataset is partitioned into $J$ subsets of size $m$ where $m\ll n$. Then, $J$ local GP experts are trained on each of these subsets, thereby reducing the training complexity to $O(Jm^3)$. 
Importantly, this approach scales to large datasets because training and prediction with each expert can be distributed across computing units~\cite{pmlr-v37-deisenroth15}. For instance, \citet{DBLP:conf/nips/RasmussenG01, 10.5555/3008751.3008843,trapp2019structured} consider mixture-of-expert models (MoEs). In particular,  \citet{trapp2019structured} propose a sum-product network with local-expert GP leaves allowing for tractable and exact posterior inference. Other approaches leverage product-of-experts models (PoEs) \cite{Tresp2000, Cao2014GeneralizedPO}, whereby a global prediction can be obtained by means of averaging the predictions of local experts.  Generalisations of these models can control the relevance of different experts when making predictions \cite{Cao2015TransductiveLO,pmlr-v37-deisenroth15,liu2018generalized}. 

In this work, we focus on PoEs because closed-form inference and training are tractable, which is not the case with typical MoEs. However, previous PoE approaches to combining predictions at test time suffer from unrealistic over- or under-estimation of the variance and erratic mean behaviours. This holds especially when the number of points $m$ assigned to each expert is low, in which case a significant number of experts are weak \cite{pmlr-v37-deisenroth15}. These approaches are thus not overly robust to variations in $m$, which is a significant shortcoming. 
Unfortunately, scalability requires the number of points per expert to be reasonably small due to the $O(m^{3})$ scaling of individual experts. We propose a solution to these problems by controlling the sparsity of expert weights through a tempered softmax at test time, leveraging tools from the the extensive uncertainty calibration literature \cite{platt1999probabilistic,bishop2012bayesian,guo2017calibration}. We also propose a novel principled PoE approach arising from the optimal transport literature, which we name the barycenter of GPs, and demonstrate that its performance is competitive to the best PoE models on small and large-scale datasets. We demonstrate empirically that calibrating expert weights lead to substantial performance gains in both mean prediction and uncertainty quantification. We also discuss common failures of PoE models extensively and propose guidelines to remediating these.

\textbf{Contributions:} $1)$ We introduce a new method for averaging GP experts based on optimal transport theory that performs competitively with the best-performing PoE models. $2)$ We propose a solution to the shortcomings of previously proposed PoEs, based on controlling the weight sparsity. $3)$ We analyse and disentangle contradictory results arising from recent GP experts papers on commonly used PoEs.

\section{Gaussian Processes}
 
Gaussian processes are powerful nonparametric Bayesian models, often used for regression. A GP is defined as a collection of random variables, every finite subset of which is jointly Gaussian distributed~\cite{Rasmussen:2005:GPM:1162254}. GPs are fully defined by a mean $m(\cdot)$ and a kernel $k(\cdot,\cdot)$. 
 
Consider a regression problem with a training dataset $\{\v{x}_i,y_i\}^n_{i=1}$ of $n$ noisy observations $y_i=f(\v{x}_i)+\epsilon$, where $\epsilon \sim\mathcal{N}(0,\sigma_y^2)$. With a GP prior on $f$, it follows that $f(\v{x}) \sim \mathcal N(\v{m}_x,\m{K}_x+\sigma_{y}^2\m{I})$ where $(\v{m}_{x})_{i}=m(\v{x}_{i})$ and $({K}_{x})_{ij}=k(\v{x}_{i},\v{x}_{j})$. The mean and variance of the Gaussian posterior predictive distribution of the function value $f(\v{x}_*)$ at a test point $\v{x}_*$, are given by
\begin{align*}
\mathbb{E}[f_*|\v{x}_*, \m{X}, \v{y}]& = \mathbf{m_{x_*}}+\m{K}_*(\m{K}_x+\sigma_{y}^2\mathbf{I})^{-1}(\mathbf{y}-\mathbf{m_x)},
\\
\text{var}[f_*|\v{x}_*, \m{X}, \v{y}]& = \m{K}_{**}-\mathbf{k}_*^{T}(\m{K}_x+\sigma_{y}^2\mathbf{I})^{-1}\mathbf{k}_*, 
\end{align*}
respectively, where $\m{K}_{**}=k(\mathbf{x}_*,\mathbf{x}_*)$ and  $\m{K}_*=k(\mathbf{X},\mathbf{x}_*)$. Here $\m{X}, \v{y}$ contain the training inputs and targets, respectively.
Kernel hyperparameters and the noise parameter $\sigma_y$ are learned by maximising the log-marginal likelihood
\begin{equation}
\label{eq:gplikelihood}
    \log p(\v{y}|\m{X},\v{\theta}) =\log \mathcal{N}(\v{y}|\v{m}_x,\m{K}_{xx}+ \sigma_y^2\mathbf{I}\big).
\end{equation}
Computing \eqref{eq:gplikelihood} requires the inversion of the matrix $\m{K}_{xx}+ \sigma_y^2\mathbf{I}\in \mathbb{R}^{n\times n}$, so that GP training scales in $O(n^{3})$, where $n$ is the size of the training dataset. Optimizing the log-marginal likelihood in \eqref{eq:gplikelihood} and the computation of the posterior predictive distribution at a test input $\v{x}_*$ become computationally intractable for large training sets. 
 
Several approaches have been explored to avoid the cubic training cost of GPs. These are mostly based on either sparse approximations and structure-exploiting assumptions to the covariance matrix \cite{10.5555/1046920.1194909,Titsias09variationallearning, Hensman:2013:GPB:3023638.3023667,pmlr-v37-wilson15} or training distributed (weak) experts on subsets of the full dataset \cite{Tresp2000, Cao2014GeneralizedPO, pmlr-v37-deisenroth15,trapp2019structured,liu2018generalized}. An alternative is to use large-scale computing infrastructure and incomplete Cholesky decompositions \cite{Wang2019}.

\subsection{Sparse Gaussian Processes}
Sparse GPs \cite{10.5555/1046920.1194909,NIPS2005_2857} leverage inducing inputs to reduce the rank of the matrix to be inverted. Sparse variational GPs extend this by introducing a variational approximation to the posterior \cite{Titsias09variationallearning}, treating inducing inputs as variational parameters, and mini-batching \cite{Hensman:2013:GPB:3023638.3023667} to scale.
\citet{pmlr-v37-wilson15} exploit structural assumptions and combine inducing-point approaches with Kronecker and Toeplitz methods to perform kernel approximations leading to increased scalability. The approximation quality of sparse GPs relies on the number of inducing points, and a large number of these can be required to represent the local structures of fast varying functions. %
  \begin{figure*}
\centering
\subfigure[PoE]{
  \includegraphics[width=0.23\hsize]{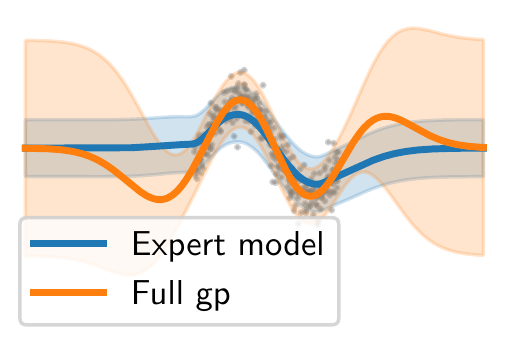}
  \label{fig:poe}
  }
  \subfigure[gPoE]{
  \includegraphics[width=0.23\hsize]{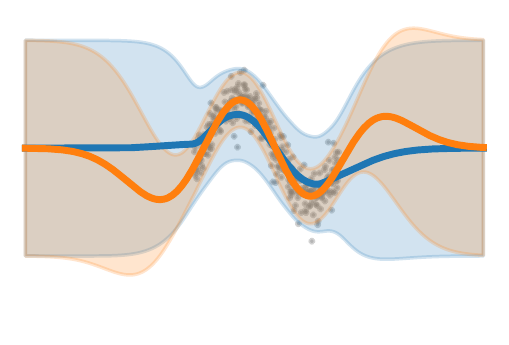}
  \label{fig:gpoe}
  }
 \subfigure[BCM]{
  \includegraphics[width=0.23\hsize]{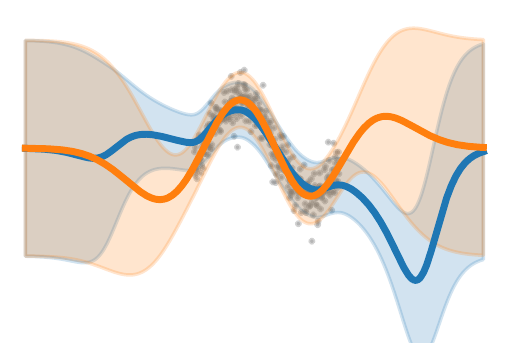}
  \label{fig:bcm}
  }
  \subfigure[rBCM]{
  \includegraphics[width=0.23\hsize]{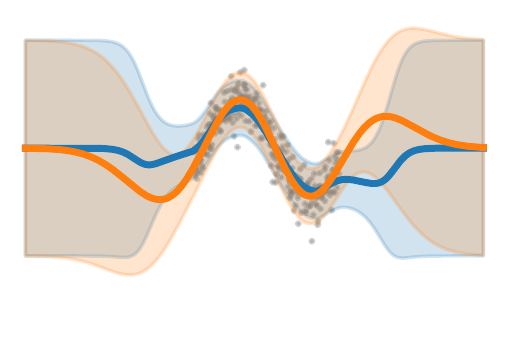}
  \label{fig:rbcm}
  }
\caption{Different expert models trained on synthetic data with three points per GP expert on a dataset of 300 observations. \subref{fig:poe} PoE; \subref{fig:gpoe} gPoE; \subref{fig:bcm} BCM; \subref{fig:rbcm} rBCM. All models display some shortcomings in their vanilla forms. For instance \subref{fig:poe}: over-confidence, \subref{fig:gpoe} under-confidence within data region, and  \subref{fig:bcm}-\subref{fig:rbcm} erratic mean in the transitioning region.}
\label{fig:mn1}
\end{figure*}
\subsection{Gaussian Process Experts}
\label{sec:experts}
 
Another approach to scaling GPs to large datasets is to use expert models. Here, multiple GPs are trained on subsets of the data, and predictions are recombined using either a  product-of-expert (log-opinion pool) approach
\cite{hinton1999products,Tresp2000,Cao2014GeneralizedPO, pmlr-v37-deisenroth15,rulliere:hal-01345959, Bertone2019}, or a mixture-of-expert (linear-opinion pool) approach \cite{10.5555/3008751.3008843,DBLP:conf/nips/RasmussenG01,trapp2019structured}. MoEs are useful in heteroskedastic and nonstationary settings, but do not typically allow tractable posterior inference, by contrast with PoEs.  

In this paper, we thus focus on product-of-expert models with $M$ experts, which all share hyperparameters. We first describe the training of such models. Assuming a full GP is the model we seek to approximate, sharing kernel hyperparameters automatically regularises the population of experts: individual experts can not overfit to the local subset of the data they are fed with due to this shared set of hyperparameters.  Assuming independence across experts (given the training data), the log-marginal likelihood is
\begin{equation}
  \log p(\v{y}|\m{X}, \v{\theta})   = \sum_{j=1}^{J} \log p_{j}(\v{y}^{(j)}|\v{x}^{(j)}, \v{\theta}),
  \label{eq:lml poe}
\end{equation}
where $\{\v{x}^{(j)},\v{y}^{(j)}\}$ is the data assigned to the $j^{th}$ expert. To train the model, we maximise the log-marginal likelihood~\eqref{eq:lml poe} with respect to the (shared) kernel hyperparameters \cite{pmlr-v37-deisenroth15}.
Training can be distributed across diverse compute clusters, enabling scaling with total time complexity $O(Jm^{3})$ where $J$ is the number of experts, and $m\ll n$ is the size of the training set of each expert. With $J$ compute nodes, the complexity per node reduces to $O(m^3)$. This is in stark contrast to the $O(n^{3})$ scaling of full GPs.  

In the following, we describe the process of predicting with product-of-GP-experts models. In particular, we introduce several approaches to recombining predictions from trained experts. We note that an important particularity of these models is that all predictive distributions $p(f_*|\v{x}_*)$ of function values are Gaussians, which is not the case with MoEs. Also, throughout the paper, aggregation is performed in function space, and the likelihood is subsequently applied.
 
\textbf{(Generalised) product of experts -- (g)PoE\quad}
The (g)PoE aggregates predictions of $M$ experts at test point ${\mathbf{x}_*}$ via
\begin{equation}
    p(f_*|\v{x}_*)\propto\prod\nolimits^{J}_{j=1}{p}_j^{\beta_j(\v{x}_*)}(f_*|\mathbf{x}_*,\mathcal{D}^{(j)}),
\end{equation}
where the predictive mean and precision are
\begin{align*}
    &m_{(g)poe}(\v{x}_*)= \sigma_{(g)poe}^{2}(\v{x}_*)\sum\nolimits_{j=1}^J\beta_{j}(\v{x}_*)\sigma_{j}^{-2}(\v{x}_*)m_{j}(\v{x}_*),\\
&\sigma_{(g)poe}^{-2}(\v{x}_*)=\sum\nolimits_{j=1}^J\beta_{j}(\v{x}_*)\sigma_{j}^{-2}(\v{x}_*),  
\end{align*}
respectively. Here,  $\mathcal{D}^{(j)}=\{\v{x}^{(j)},\v{y}^{(j)}\}$ is the data assigned to expert $j$,  $\beta_j(\v{x}_*)$ controls the contribution of expert $j$ at  $\v{x}_*$ (typically a measure of its confidence at $\v{x}_*$), and the PoE model is recovered when setting $\beta_j(\v{x}_*) = 1$ for all $j$.  As the number of experts $J$  increases, the PoE's aggregated variance vanishes, which leads to overconfident predictions \cite{pmlr-v37-deisenroth15,liu2018generalized}. An illustration of such behaviour is shown in Figure \ref{fig:poe}. 

The gPoE with uniform weights $\sum_{j}\beta_{j}(\v{x}_*)=1$ falls back to the prior far from training points, which is a desirable property. However, a drawback is that it over-estimates the variance close to training points~\cite{pmlr-v37-deisenroth15} when setting the weights uniformly ($\beta_{j}(\v{x}_*)=\frac{1}{J}$). We also observe such behaviour in Figure \ref{fig:gpoe}.
 
\textbf{(Robust) Bayesian committee machine -- (r)BCM}
\quad The (robust) Bayesian committee machine (r)BCM~\cite{Tresp2000, pmlr-v37-deisenroth15} assumes conditional independence $\mathcal{D}_{i} \indep \mathcal{D}_{j}|f_{*}$. By repeated application of Bayes' theorem, we obtain the predictive distribution 
\begin{align}
    p(f_*|\v{x}_*)=\frac{\prod\nolimits^{J}_{j=1}{p_j}^{\beta_j(\v{x}_*)}(f_*|\mathbf{x}_*,\mathcal{D}^{(j)})}{ p^{-1+\sum_{j}\beta_j(\v{x}_*)}(f_*|\mathbf{x}_*)}
    \label{eq:rBCM model}
\end{align}
at test point ${\mathbf{x}_*}$. Then the predictive mean and precision are
\begin{align*}
    &m_{(r)bcm}(\v{x}_*)= \sigma_{(r)bcm}^2(\v{x}_*)\sum_{j=1}^J\beta_{j}(\v{x}_*)\sigma_{j}^{-2}(\v{x}_*)m_{j}(\v{x}_*), \\
    &\sigma_{(r)bcm}^{-2}(\v{x}_*)=\sum_{j=1}^J\beta_j(\v{x}_*)(\sigma_{j}^{-2}(\v{x}_*)-\sigma_*^{-2})+\sigma_*^{-2},
\end{align*}
respectively. BCM is recovered when $\beta_j(\v{x}_*)=1 $ for all $j$. This predictive distribution guarantees that the model falls back to the prior far from training data.  However, the BCM exhibits uncharacteristic behaviour in regions transitioning from high to low-density data~\cite{pmlr-v37-deisenroth15}; see Figure \ref{fig:bcm}. The rBCM mitigates some of the issues of the BCM and allows for flexible weighting of GP experts, via $\beta_j(\v{x}_*)$, but it still exhibits problematic behaviour in regions with density transitioning; see Figure \ref{fig:rbcm}.
 
\paragraph{Generalised rBCM -- grBCM \quad}\citet{liu2018generalized} proposed the grBCM in which a master expert communicates with the children experts leading to a consistent predictive distribution of the form
\begin{align}
    p(y_*|\v{x}_*)=\frac{\prod\nolimits^{J}_{j=2}{p}_{+j}^{\beta_j(\v{x}_*)}(y_*|\mathbf{x}_*,\mathcal{D}^{(+j)})}{ p^{-1+\sum_{j=2}^{p}\beta_j(\v{x}_*)}(\mathcal{D}^{c}|y_*,\mathbf{x}_*)}.
    \label{eq:grBCM model}
\end{align}
$\mathcal{D}^c$ is the global data assigned to the master expert, and $\mathcal{D}^{(+j)}=\{\mathcal{D}^c,\mathcal{D}^{(j)}\}$ is the data of the $j^{th}$ expert aggregated with the data of the master expert. The predictive mean and variance of \eqref{eq:grBCM model} are
\begin{align*}
    m_{grb}(\v{x}_*)&= \sigma_{grb}^2(\v{x}_*)\Big[\sum_{j=2}^J\beta_{j}(\v{x}_*)\sigma_{+j}^{-2}(\v{x}_*)m_{+j}(\v{x}_*)\nonumber\\
    &\quad  -(\sum_{j=2}^J\beta_j-1)\sigma_c^{-2}(\v{x}_*)\mu_c(\v{x}_*)\Big], \\
    \sigma_{grb}^{-2}(\v{x}_*)&=\sum_{j=2}^J\beta_j(\v{x}_*)(\sigma_{+j}^{-2}(\v{x}_*)-\sigma_c^{-2}(\v{x}_*))+\sigma_c^{-2}(\v{x}_*),
\end{align*}
where $m_{+j}(\v{x}_*)$ and $\sigma_{+j}^2(\v{x}_*)$ are the predictive mean and variance of the $j^{th}$ expert at $\v{x}_*$, conditioned on the aggregated dataset $\mathcal{D}^{(+j)}$.
\citet{liu2018generalized} perform aggregation in $y$-space, in contrast to \citet{pmlr-v37-deisenroth15}, who perform it in $f$-space. The former is not directly applicable in non-conjugate cases (e.g., classification), and can lead to erratic mean and variance behaviours (especially when used for rBCM/BCM/PoE as observed in \cite{liu2018generalized,JMLR:v20:18-374}). We therefore consider in this paper aggregation in $f$-space and discuss differences between both approaches further in later sections.
\paragraph{Likelihoods}
As expert averaging is performed in function space throughout the paper, we will need to map the aggregated predictive GP distribution $p(f_*)$ through a likelihood function to predict labels $y_*$. In the conjugate regression case with a Gaussian likelihood, this can be done in closed form~\cite{Rasmussen:2005:GPM:1162254}.
For classification, we consider non-conjugate likelihoods, such as the Bernoulli or Poisson likelihoods. Since the aggregated predictive distribution $p(f_*)$ in PoEs is Gaussian, we obtain the expected predicted label by averaging under the posterior predictive latent distribution
\begin{align} \label{eq:gpclass} 
\mathbb{E}[y_*|\v{x}_*]= \int \phi(f(\v{x}_*)) \mathcal N(f_* | m(\v{x}_*), \sigma^2(\v{x}_*)) df_*,
\end{align}
where $\phi$ is a classification likelihood (e.g., Bernoulli, Probit). The integral in~\eqref{eq:gpclass} is intractable, but we can resort to standard approximate inference techniques for GP classification, such as MAP estimation, Laplace approximation, expectation propagation, variational inference, or numerical integration~\cite{Rasmussen:2005:GPM:1162254, pmlr-v38-hensman15}. Similarly, the marginal likelihood, which we use for training the experts, becomes intractable. Therefore, we use stochastic variational inference to train models in that setting \cite{pmlr-v38-hensman15}, and apply the same strategies for training and prediction with other GP expert models.

\section{Barycenters of Predictive Distributions}
\label{sec:barycenter}
 
Now, we propose a new way of combining experts' predictions leveraging optimal transport theory. We begin by introducing two important tools, namely the Wasserstein distance and barycenter between 1D Gaussians, noting that  both can be computed using simple closed-form formulas.

 Given two Gaussians $\mu = \mathcal{N}(\v{m}_1, \v{K}_1)$ and  $\nu = \mathcal{N}(\v{m}_2, \v{K}_2)$, we define the 2-Wasserstein distance between them as \cite{Villani2008OptimalTO}
 \begin{align}
 \label{eq:wass}
     \mathcal{W}_2^{2}(\mu,\nu) &= \Vert\v{m}_1-\v{m}_2\Vert^2_2\nonumber\\
     &\quad +Tr\Big(\v{K}_1 + \v{K}_2 -2 (\v{K}_1^{\frac{1}{2}}\v{K}_2\v{K}_1^{\frac{1}{2}})^{\frac{1}{2}}\Big).
 \end{align}
  Equation~\eqref{eq:wass} can be interpreted as the minimal expected cost of transporting mass from the Gaussian $\mu$ to the Gaussian $\nu$. 
 
 Given that distance, the barycenter between Gaussian-distributed $\mu_{1},...,\mu_J$  with weights $\v{\beta}$ is
 \begin{equation}
 \label{eq:bar}
     \bar{\mu} = \argmin_{\mu}\sum\nolimits_{j=1}^{J}\beta_{j}\mathcal{W}^{2}_{2}(\mu_{j},\mu),
 \end{equation}
where $\sum_{j}\beta_{j}=1$, $0\leq \beta_{j} \leq 1 $. \citet{pub.1023863441} show that if $\mu_{j} = \mathcal N(\v{m}_{j},\v{K}_{j})$ for all $j$, the Wasserstein barycenter with weights $\v{\beta}$ is itself a Gaussian measure $\bar{\mu}=\mathcal N(\bar{\v{m}}, \bar{\m{K}})$, where 
\begin{equation}
\label{eq:gaussbar}
    \bar{\v{m}} = \sum_{j=1}^{J}\beta_{j} \v{m}_{j}, \quad
    \bar{\m{K}}=\sum_{j=1}^J\beta_{j}(\bar{\m{K}}^{\frac{1}{2}}\m{K}_{j}\bar{\m{K}}^{\frac{1}{2}})^{\frac{1}{2}}.
\end{equation}
The authors also propose a fixed-point iteration algorithm to efficiently compute $\bar{\m{K}}$ in~\eqref{eq:gaussbar}.
\begin{figure}
    \centering
    \includegraphics[width=0.68\hsize]{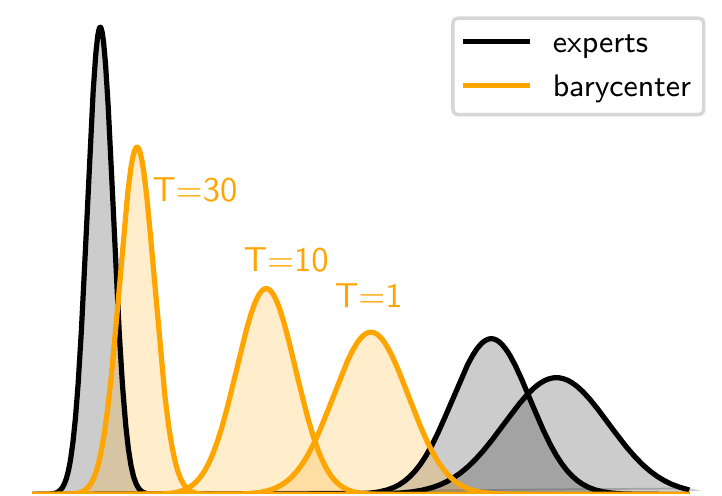}
    \caption{Illustration of the barycenter of GPs with tempered softmax weighting. At $x_*$, one expert (left) is highly confident about its prediction, and two are highly unconfident (right). As temperature increases, only confident experts get weight (sparsity increases), thus the barycenter is pulled towards the confident expert.}
    \label{fig:bary_illus}
\end{figure}

In the following, we discuss our approach to aggregating GP experts' predictions for regression and classification. In all product-of-experts models we discussed, each expert computes a predictive distribution of the form $p_{j}(f(\v{x}_*)|\mathcal{D}^{(j)}) = \mathcal N(m_{j}(\v{x}_*),\sigma_{j}^{2}(\v{x}_*))$, where $m_{j}$ and $\sigma_{j}^{2}$ are the posterior predictive mean and variance of the $j${th} GP expert at test point $\v{x}_*$. Since these distributions (in latent space of $f$) are all Gaussian (by definition of the GP), we propose combining these into their weighted 2-Wasserstein barycenter using \eqref{eq:gaussbar}, which can be computed in closed form in the one-dimensional case \cite{bonneel2013sliced}. 
We obtain the closed-form Gaussian predictive distribution
\begin{align}
    p(f_*|\v{x}_*)&=\mathcal N(m_{bar}(\v{x}_*), \sigma_{bar}^2(\v{x}_*))\label{eq:barycenter average}\\
    m_{bar}(\v{x}_*) &= \sum_{j=1}^{J}\beta_{j}(\v{x}_{*}) m_{j}(\v{x}_{*}),\\
    \sigma_{bar}^2(\v{x}_*) &=\sum_{j=1}^J\beta_{j}(\v{x}_{*})\sigma_{j}^{2}(\v{x}_{*}).\label{eq:meanvarbar}
    \end{align}
The barycenter of GPs is a product-of-experts variant, and the mean and variance of the predictive distribution consist of the weighted average of predictive means and variances of the experts. Importantly, such weights can be a function of test points, analogously to the gPoE and the rBCM. 

We train the barycenter of GPs following the training procedure of other PoEs discussed in Section \ref{sec:experts}, namely by optimising the marginal likelihood \eqref{eq:lml poe}, and we share expert hyperparameters for regularising the expert pool. 

The barycenter of GP's predictive distribution is deeply connected to that of previously proposed PoEs. In particular, the aggregated mean is a weighted mean of the experts' predictive means, which is also the case for other expert models. The aggregated variance is a weighted mean of experts' variances, which has a similar interpretation to the predictive precision of other PoEs, itself a weighted mean of the experts' precisions. The barycenter of GPs falls back to the prior outside the data regime which is a highly desirable property, and is also true for gPoE with uniform weights, and rBCM. Further connections are discussed in Section \ref{sec:fixing}.

WASP \citep{Srivastava2015} leverages a related idea, which consists in averaging subset posteriors using Wasserstein barycenters. However, they average discrete measures consisting of samples from the different posteriors at a discretizsed set of points, and then have to solve a large linear problem to compute the barycenter. By contrast, we average marginal posterior predictive distributions, which is done in closed-form leveraging the known closed-form of barycenters of Gaussians in 1D.

%
 
%

\section{Calibrating Product-of-Experts}
\label{sec:fixing}
%
 In the previous sections, we introduced several approaches to combining predictions of local GP experts, including our proposal, the barycenter of GPs. We also discussed shortcomings of previous PoE approaches in low-data regimes, including under- (Figure~\ref{fig:poe}) and over-estimation of the variance (Figure~\ref{fig:gpoe}), but also erratic and uncharacteristic behaviours of the mean and variance predictions (Figures~\ref{fig:bcm}--\ref{fig:rbcm}).
 These behaviours are exacerbated when the number of points assigned per expert is low, which leads to a significant number of weak experts\footnote{We refer to weak experts as experts that provide calibrated predictions only on local subsets of the data manifold.}. 
 
 Whilst exact Gaussian processes are well-known for well-calibrated uncertainty estimates, approximate Bayesian methods fall prey to inferior calibration. These issues in the context of sparse GP approximations are discussed in depth by \citet{NIPS2016_6477}. Our aim in this section is to remediate such calibration issues for PoE models. 
 There has been a significant recent emphasis on uncertainty calibration in the deep learning community \cite{guo2017calibration}, and we will extend tools from this literature to the problem of training product-of-experts-based GP approximations.   

%
 


 \begin{figure*}
\centering
  \subfigure[diff entr, 100 pts/exp]{
  \includegraphics[width=0.23\hsize]{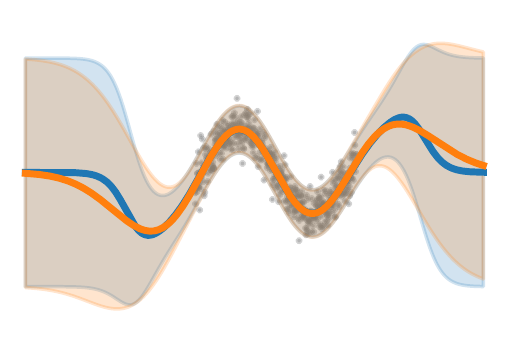}
  \label{fig:unif200}
  }
  \subfigure[diff entr, 20 pts/exp]{
  \includegraphics[width=0.23\hsize]{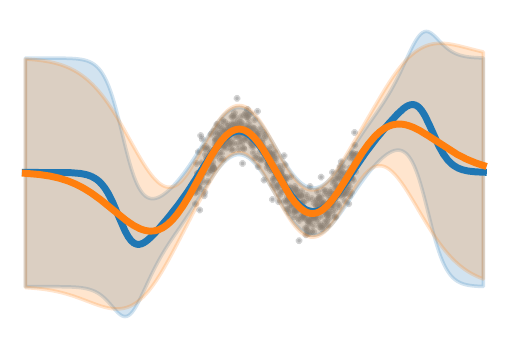}
  \label{fig:unif10}
  }
 \subfigure[diff entr, 4 pts/exp]{
  \includegraphics[width=0.23\hsize]{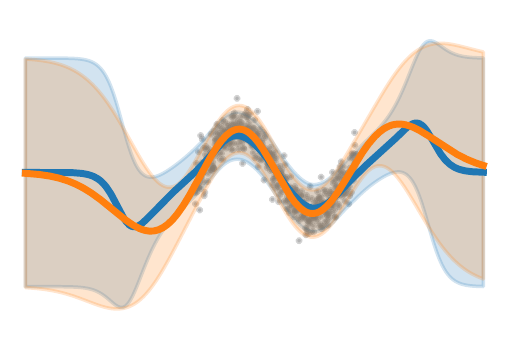}
  \label{fig:unif4}
  }
  \subfigure[diff entr, 2 pts/exp]{
  \includegraphics[width=0.23\hsize]{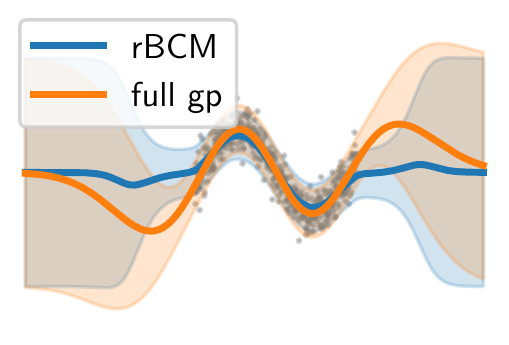}
  \label{fig:unif2}
  
  }\\
  \subfigure[soft-var, 100pts/exp]{
  \includegraphics[width=0.23\hsize]{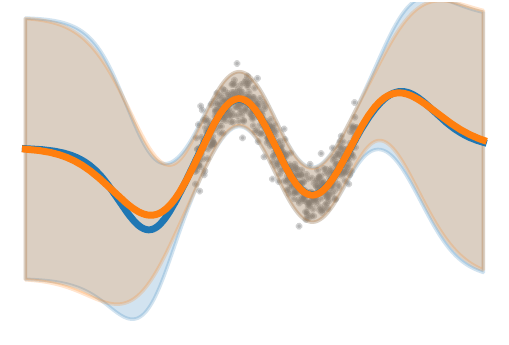}
  \label{fig:soft200}
  }
  \subfigure[soft-var, 20 pts/exp]{
  \includegraphics[width=0.23\hsize]{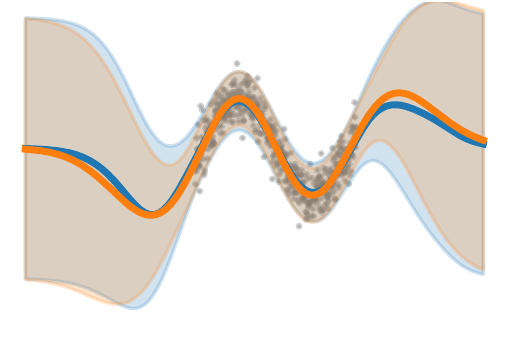}
  \label{fig:soft10}
  }
 \subfigure[soft-var, 4 pts/exp]{
  \includegraphics[width=0.23\hsize]{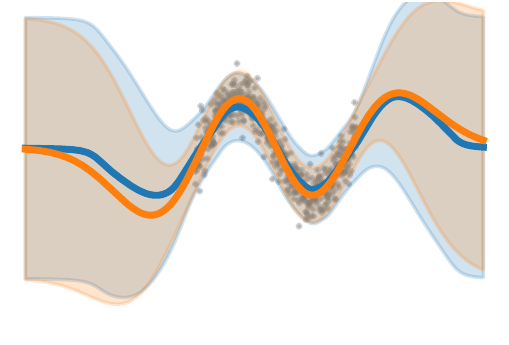}
  \label{fig:soft4}
  }
  \subfigure[soft-var, 2 pts/exp]{
  \includegraphics[width=0.23\hsize]{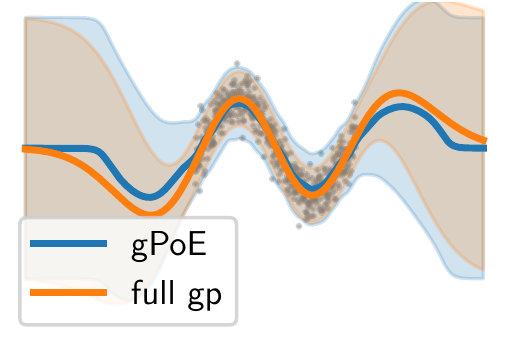}
  \label{fig:soft2}
  }
\caption{Full GP baseline (orange) and expert models (blue) trained on synthetic data with a decreasing number of points per experts (Left to Right), and for different weighting methods: rBCM with differential entropy in Figures \subref{fig:unif200}--\subref{fig:unif2} and the gPoE with proposed softmax-variance in Figures \subref{fig:soft200}--\subref{fig:soft2}. Our method is significantly more robust to variations in the number of points per experts.
}
\label{fig:robustness_ppe}
\end{figure*}
The prevalence of weak experts is significantly affected by the data assignment strategy. For example, when using stationary kernels, clustering-based partition approaches tend to create localised experts which leads to greater weak expert prevalence. The latter approach is intuitively sensible if we choose stationary kernels, as expert approaches can be interpreted as divide-and-conquer strategies.  However, this strategy can have disastrous consequences if expert weights are not properly regulated. Indeed, the lower the number of points per expert, the weaker the experts are overall if the training data associated with these experts is not dense in the vicinity of test inputs. This can be observed in Figure \ref{fig:robustness_ppe} (Top), where pathologies arise as the number of points per expert decreases significantly. This is mainly caused by the poorly regulated expert weighting. 

In that setting, weight sparsity has to increase to alleviate the weakness of most experts by relying only on locally-calibrated predictions. 
In the following, we propose a solution to such shortcomings that can be applied to gPoE, rBCM and the barycenter of GPs. 

The softmax function provides a natural mechanism for controlling the sparsity of experts' importance weights. In particular, an (inverse) temperature parameters $T$ can directly control the degree of smoothness and sparsity in the resulting weights. Using a temperature-endowed softmax to combat miscalibrated predictions has seen widespread use, ranging from hierarchical mixtures of experts \cite{bishop2012bayesian} to support vector machines \cite{platt1999probabilistic} and deep learning \cite{guo2017calibration}.
 
We adapt these ideas to weighted ensembles of GP experts, such as the gPoE, the rBCM and the barycenter. We therefore propose a general expression for expert weights as
  \begin{equation}
     \beta_{j}(\v{x}_*)\propto\exp({-T\, \psi_j(\v{x}_*)}),\quad \sum_{j=1}^M \beta_j(\v{x}_*) = 1,
     \label{eq:softmax}
 \end{equation}
 where $T$ is an (inverse) temperature parameter that controls the sparsity between experts by multiplicatively compounding the weights of stronger experts. The functional $\psi_j(\v{x}_*)$ describes the level of confidence of the $j$th expert at test point $\v{x}_*$.  We provide an illustration of such framework in Figure \ref{fig:bary_illus}. In particular, we plot the barycenter of GP experts' predictive distribution at $\v{x}_*$ under several temperature values, highlighting that as temperature increases, the barycenter gets pulled towards the most confident expert, i.e., uncertain experts are not given weight in the prediction.
 
 We now discuss the choice of confidence functional $\psi$. We set $\psi_j(\v{x}_*)$ to the posterior predictive variance at $\v{x}_*$, i.e.,
   \begin{equation}
     \psi_{j}(\v{x}_*)=\sigma_{j}^{2}(\v{x}_*).
 \end{equation}
 Intuitively, this will give high weight to experts with low posterior predictive variance (high confidence) in their prediction. Such experts have training data close to test points (all experts share the same hyperparameters), and should thus have a high contribution in the final prediction. Our proposal can also be combined with the previously proposed differential entropy weighting \cite{Cao2014GeneralizedPO}
 \begin{align}
 \psi_j(\v{x}_*)=\tfrac{1}{2}(\log\sigma_*^{2}-\log\sigma_{j}^{2}(\v{x}_*))
 \end{align}
 or with the Wasserstein distance \eqref{eq:wass}, leveraging its closed-form computation in the 1D case \cite{pub.1023863441}, which has the same complexity as differential entropy. 
 In the infinite temperature limit, weight sparsity is maximised. We show that in this regime, the gPoE, the rBCM and the barycenter of GPs are equivalent, and provide proofs in Appendix \ref{sec:apdxtheory}
 \begin{proposition}
 \label{prop:equivtemp}
 In the infinite-temperature limit $T\rightarrow \infty$, and if $\psi_j=\sigma_j^2(\v{x}_*)$, the gPoE, the rBCM and the barycenter of GPs have equivalent predictive distributions.
 \end{proposition}
Intuitively, in such a regime, only the most confident experts have (equal) weight, and as a result the inverse of the weighted sum of precisions of the two former equals the weighted sum of the variances, and thus predictive distributions are equal. Under weaker assumptions, the rBCM and the gPoE are equivalent:

\begin{proposition}
\label{prop:equivpoe}
 If $\sum_j \beta_j(\v{x}_*)=1$ for all $\v{x}_*$, then $m_{rbcm}(\v{x}_*)=m_{gpoe}(\v{x}_*)$  and $\sigma^2_{rbcm}(\v{x}_*)=\sigma^2_{gpoe}(\v{x}_*)$.
\end{proposition}

Proposition \ref{prop:equivpoe} highlights that under normalised weights, gPoE and rBCM are equivalent. Therefore, under our weighting proposal, which consists of using normalised tempered softmax functionals, gPoE and rBCM's predictive distributions are equal.
\begin{table*}
\resizebox{\linewidth}{!}{
\begin{tabular}{rrrrrrrrrrrr}
\toprule
    dataset &      N &   D &     rBCM/gPoE\_unif  &           rBCM/gPoE\_var &                  rBCM\_entr                           &        BAR\_var & grBCM\_$f$  &              $\text{SVGP}_{500}$ &             linear  & full GP \\
\midrule
Concrete  &   1030 &   8 &  0.506 (\blue{0.370})  &   \textbf{0.288 (\blue{0.342})}    &   0.292 (\textbf{\blue{0.343}}) &  \textbf{0.288 (\blue{0.342})} & \textbf{0.285} (\textbf{\blue{0.339}}) &   \textbf{0.289 (\blue{0.338})} &  0.953 (\blue{0.626}) & \textbf{0.261 (\blue{0.330})}\\
Airfoil    &   1503 &   5 &  0.699 (\blue{0.474})   &   \textbf{0.411 (\blue{0.350})}    &  \textbf{ 0.409 (\blue{0.360})} &  \textbf{0.411 (\blue{0.351})} &\textbf{0.413} (\blue{\textbf{0.350}}) &    \textbf{0.409 (\blue{0.353})} &  1.096 (\blue{0.721})& \textbf{0.358 (\blue{0.331})} \\
Parkinsons &   5875 &  20 &  1.057 (\blue{0.713})   &  \textbf{0.101 (\blue{0.338})}      &   0.157 (\textbf{\blue{0.339}}) & \textbf{0.100 (\blue{0.337}) }& 0.145 (\blue{0.345})&   0.554 (\blue{0.412})           &  1.282 (\blue{0.871})& \textbf{0.079 (\blue{0.320})}\\
Power    &   9568 &   4 &  0.303 (\blue{0.318})    &       \textbf{-0.084 (\blue{0.222})} &  \textbf{-0.079 (\blue{0.223})} &  \textbf{-0.076} (\textbf{\blue{0.224}})   &  \textbf{-0.074} (\blue{\textbf{0.224}})  &  -0.044 (\blue{0.231})            &  0.098 (\blue{0.267}) & \textbf{-0.079 (\blue{0.223})}\\
Kin40K    &  40000 &   8 &  1.078 (\blue{0.693})  &      -0.329(\blue{0.186})            &   0.359 (\blue{0.191})          &  -0.339  (\blue{0.183}) &\textbf{-0.432  (\blue{\textbf{0.150}})}&     0.124 (\blue{0.263}).           &  1.419 (\blue{1.000}) &N/A\\
Protein   &  45730 &   9 &  1.379 (\blue{0.961})   &     \textbf{0.775 (\blue{0.582})}   &   0.799 (\blue{0.608})           &  \textbf{0.775 (\blue{0.583})}  &0.797 (\blue{0.607})  &    1.083 (\blue{0.715})           &  1.257 (\blue{0.850}) &N/A\\
Airline   &  \airlinesize &   7 & 1.440 (\blue{0.996})  &    \textbf{1.318 (\blue{0.908})}     &  \textbf{ 1.312} (\textbf{\blue{0.902}})           & \textbf{1.318 (\blue{0.908})}    &\textbf{1.311 (\blue{0.902})}&    1.335(\blue{0.921})  &  1.388 (\blue{0.969}) & N/A\\
    
        average &        &     &        0.923 (\blue{0.646}) &   \textbf{0.354 (\blue{0.418})}        &  0.464 (\blue{0.423})     &          \textbf{0.353} (\textbf{\blue{0.419}}) &  \textbf{0.350 (\blue{0.417})}  &             0.537 (\blue{0.461}) &         1.070 (\blue{0.758}) & N/A\\
\bottomrule
\end{tabular}
}
\caption{Average NLPD (\blue{RMSE}) for small (1K+ points) and large-scale (40K+ points) benchmarks under clustering partitioning.}
\label{Tab:nlpd-rmse}
\end{table*}
  
 
\section{Experiments}
\label{sec:exp}
Throughout this section, we evaluate the performance of our approaches to calibrating GP experts when applied to regression and classification, while comparing with sparse variational methods and previous approaches to local-expert weighting and averaging. We consider performance metrics including the negative log-predictive density (NLPD), and the root mean squared error (RMSE).
\footnote{Code available at \url{https://github.com/samcohen16/Healing-POEs-ICML}}\footnote{Datasets are from \url{https://github.com/hughsalimbeni/bayesian_benchmarks}.}

 \textbf{Baselines:} We consider the gPoE, rBCM and barGP with random  and $K$-means partitioning to assess the effect of the data assignment strategy. For the rBCM, gPoE and barGP, we evaluate the proposed softmax weighting strategy (BAR\_var, rBCM\_var, gPoE\_var) with different temperature choices as proposed in Section \ref{sec:fixing}. We also evaluate differential entropy (\_entr) weighting \cite{Cao2014GeneralizedPO}  and uniform weighting (\_unif). According to Remark \ref{prop:equivpoe}, when using normalised weights, the gPoE and rBCM are equivalent. We thus combine their results into `rBCM/gPoE\_...'. We further compare to the grBCM \cite{liu2018generalized} and showcase results of this model obtained by averaging in $y$-space as proposed in \cite{liu2018generalized}, and in $f$-space. For all other expert models, averaging is done in $f$-space following \cite{pmlr-v37-deisenroth15}.  Finally, we also consider a full GP baseline and linear regression.
 %
 
%
 
\subsection{Regression}

We evaluate the performance of our approach to setting local experts' weights and compare it to previous weighting methods. 
In particular, we evaluate the robustness of the rBCM using differential entropic weighting as motivated by \citet{pmlr-v37-deisenroth15}, and the gPoE and barycenter with softmax-variance weighting (proposed in this paper), when reducing the number of points per experts. As motivated in Section~\ref{sec:fixing}, the softmax weighting should encourage expert sparsity, and as such be effective when the number of points per experts decreases (causing the number of strong experts to decrease). In this case, we set the temperature $T$ to $15$ (for $T\geq15$, sparsity is well-controlled; see Figure \ref{fig:rob_park}).

Figure \ref{fig:robustness_ppe} shows that the gPoE with softmax-variance weighting provides sensible and calibrated predictions even with only two points per experts, while the rBCM with differential entropic weights leads to erratic mean and variance behaviours in the transitioning region even with $20$ points per experts. Thus, encouraging sparsity in the expert weights through the variance-softmax weighting enables expert models to be robust to the reduction in the number of points per experts, thereby addressing a shortcoming of local-expert models. Also, the erratic behaviour in the transitioning region appears remediated. With very weak experts, it is unrealistic to expect uncertainties that are identical to the full GP’s uncertainty. Importantly, the predictions are (moderately) on the conservative side for the softmax-variance weighting, which is preferable to overconfidence.  We report similar behaviours for the barycenter combination (Section~\ref{sec:barycenter}) in the Appendix (Figure \ref{fig:robustness_bar}).

We now perform a large-scale evaluation of the different expert models with different choices of weighting, including our approach (softmax-variance) and previous approaches (uniform for gPoE and differential entropy for rBCM) on 7 datasets of size ranging between $1000$ and $\airlinesize$. We also use  $\text{SVGP}_{500}$, grBCM, and linear regression baselines. For softmax weightings, we use a temperature of $100$, which performs well across small and large-scale benchmarks (i.e., it induces enough weight sparsity). We provide extensive additional results with different temperatures and random partitioning in the Appendix.

Table \ref{Tab:nlpd-rmse} shows that the gPoE and the barGP with softmax-variance weighting perform on par with grBCM and outperform all other models on all datasets. They significantly outperform the SVGP with $500$ inducing points, but also the rBCM with differential entropy weighting, and linear regression across benchmarks. Moreover, the gPoE with softmax-variance weighting outperforms the gPoE with uniform weighting by a large margin across small and large-scale datasets. Also, whilst the grBCM outperforms rBCM with differential entropic weights across datasets, our weighting proposal (softmax-variance) outperforms or performs on par with it on all datasets, besides on Kin40k, while having a significantly lower prediction cost.

This demonstrates that controlling the sparsity of expert weights heals issues of the product-of-expert models and leads to more calibrated uncertainty quantification and mean estimation, while having the same running cost (and a significantly lower prediction cost than grBCM). 
The complexity of predictions of the grBCM is $8\times$ higher than under other expert models, because children datasets are aggregated with the master's. Therefore, these performance gains are accompanied by computational gains.

 Finally,  \citet{liu2018generalized} and \citet{JMLR:v20:18-374} found that the rBCM and the gPoE under-perform when averaging in $y$-space, which is the reason we average in $f$-space in this paper. We analyse the performance of grBCM in both regimes, and observe that the latter leads to substantial performance gains, thus motivating averaging in $f$-space for all product-of-expert models (see Tables \ref{Tab:nlpd-rmse} and \ref{tab:yavergrbcm}). We provide a more thorough discussion in Section \ref{sec:discussion}. 


 \subsection{Sensitivity and Robustness Analysis}
\label{sec:sensitivity}
\begin{figure}
\centering
\includegraphics[width=0.46\hsize]{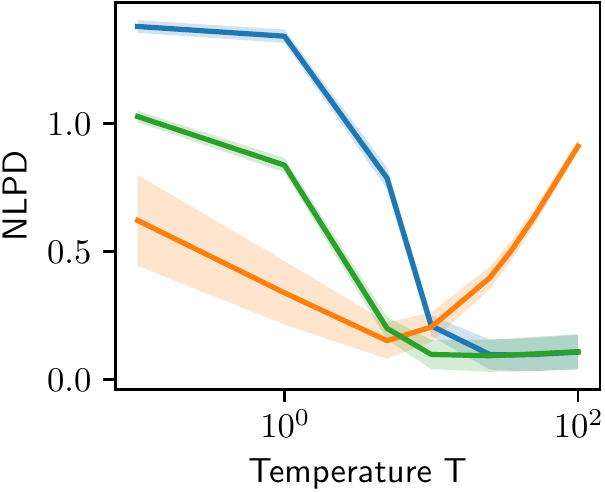}
\includegraphics[width=0.44\hsize]{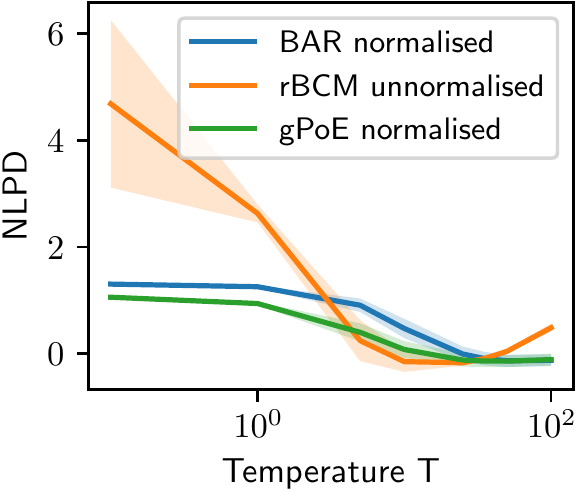}
\includegraphics[width=0.46\hsize]{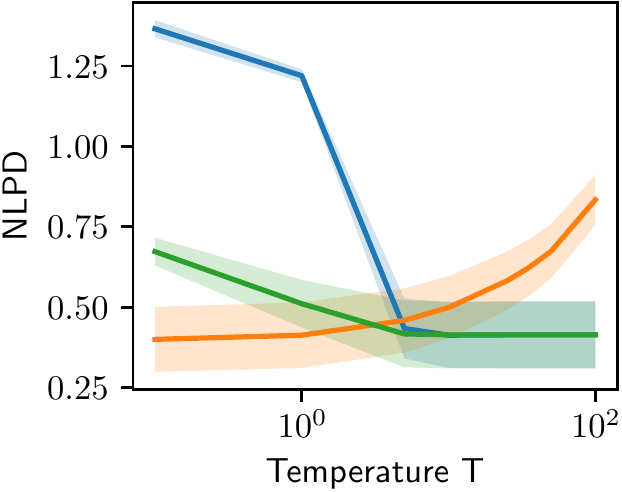}
\includegraphics[width=0.46\hsize]{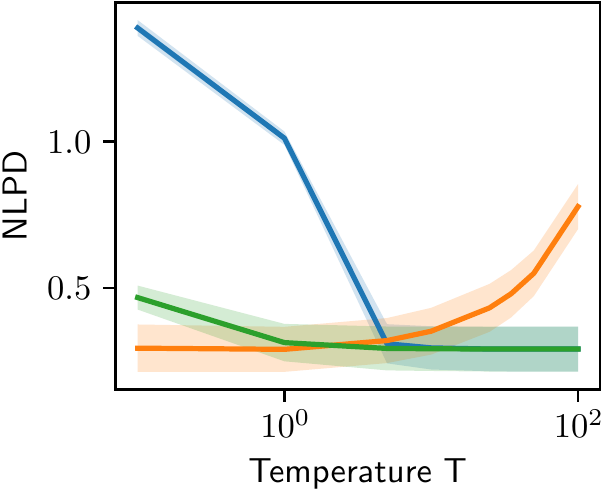}
  \caption{NLPD against temperature for different expert models with softmax-variance weighting on Parkinson (top left), Kin40K (top right), Airfoil (bottom left) and Concrete (bottom right).}
  \label{fig:rob_park}
\end{figure}

We now consider the sensitivity of the gPoE, rBCM and barGP with softmax-variance weighting to the temperature hyperparameter $T$. For the gPoE and barGP, we use the normalised version of the softmax (in which case the gPoE is equivalent to the rBCM with such weights). We also evaluate the rBCM's robustness, when using unnormalised softmax-variance weights.   To that end,  we consider  the Parkinsons, Kin40k, Airfoil and Concrete datasets, and plot the NLPD  as a function of the temperature (Figure \ref{fig:rob_park}).  We observe that the NLPD decreases monotonically until stabilising for both the gPoE and the barGP, demonstrating the robustness of these models with respect to the choice of the temperature parameter. Hence, the NLPD is stable across temperatures (for $T>15$) when using normalised weights. We also produce such an analysis for unnormalised softmax-variance weights (in which case the rBCM is not equivalent to the gPoE). In this case, the model is more sensitive to the change in temperature, and it is difficult to find a single softmax scaling that performs well across small- and large-scale benchmarks. Hence, using normalised softmax weights is important to obtaining models that are robust to the choice of temperature. 
 \begin{table*}
  \centering
\scalebox{1}{
\begin{tabular}{lllllll}
\toprule
 &   BAR\_var  &    gPoE\_unif & gPoE\_var  &    rBCM\_entr   &        $\text{SVGP}_{500}$
 \\
\midrule
Top-1-accur. &  \textbf{0.912} &  0.893 &	\textbf{0.911}	&0.895		&0.862   \\
Top-2-accur. &   \textbf{0.965} &  0.955& \textbf{0.965} &	0.956	&	0.940    \\
Top-3-accur. &  \textbf{0.982} & 0.975	&\textbf{0.982}& 0.976 &	0.967      \\
NLPD   &   \textbf{0.308} &0.382	& \textbf{0.309}
&	0.878  &	0.469        \\
\bottomrule
\end{tabular}
}
\caption{Top-$n$ accuracy and NLPDs on the MNIST dataset (PCA features).}
\label{tab:class_result}
\end{table*}

\subsection{Classification Benchmarks}
\label{sec:classification_exp}

We now assess the classification performance of expert models in a non-conjugate  multi-class classification setting (MNIST dataset).  The dataset comprises $10$ classes with a training/test split of $60,000/10,000$ images. We reduce the dimensionality  of images with PCA ($20$ principal components). Note that the overall accuracy  resulting from PCA features will not outperform the state of the art. However, PCA features provide a deterministically reproducible basis for relative comparison of various methods. We assign $500$ training points to each SVGP expert, and provide them with $100$ trainable inducing inputs each. We use a multiclass likelihood with a robust-max link function. Note that in the setting of \citet{liu2018generalized}, classification is not directly applicable because averaging is happening in $y$-space, which is more challenging in non-conjugate settings.

Table \ref{tab:class_result} shows classification results. We report top-$n$ accuracy and NLPD. Consistently, we observe that the BAR\_var and gPoE\_var  outperform all products of experts and SVGP baseline models. The difference in performance between the rBCM\_entr and gPoE\_var shows that introducing weight sparsity via a tempered softmax improves the performance as it only allows confident experts to contribute to the aggregated predictions. We observe similar performance gaps between gPoE\_unif and our proposals which suggests that using tempered softmax-variance weighting results in more informed posterior predictive means and variances.  

The improvement of the $\text{SVGP}_{100}$ expert models over a single (full) $\text{SVGP}_{500}$ is not surprising since every single SVGP expert has the modelling capacity of the global SVGP, so that the distributed models effectively work with $M$ times as many inducing inputs as the SVGP.
This suggests that the combination of sparse GPs and expert models can be useful in settings, where a large number of inducing inputs  for a full SVGP is required for good modelling.

\section{Discussion}
\label{sec:discussion}
In light of empirical results, we aim to explain our findings and compare them to those of similar papers, in particular \cite{liu2018generalized}, \cite{trapp2019structured} and \cite{JMLR:v20:18-374}. 
All three papers reported poor NLPDs and RMSEs for the PoE, the BCM, but also the gPoE and the rBCM across small and large-scale benchmarks. At first sight, this seems to contradict the empirical observations drawn in this paper. We thus aim to disentangle such conflicting conclusions. The main reason for the discrepancies is that those papers depart from the framework of \citet{pmlr-v37-deisenroth15} and aggregate GP predictions in $y$-space instead of $f$-space. However, this is problematic for such models, especially with unnormalised weights and random partitioning, because the variance shrinks for all such models, explaining the bad NLPDs reported in all three papers. 
\begin{figure}
\centering
\includegraphics[width=0.8\hsize]{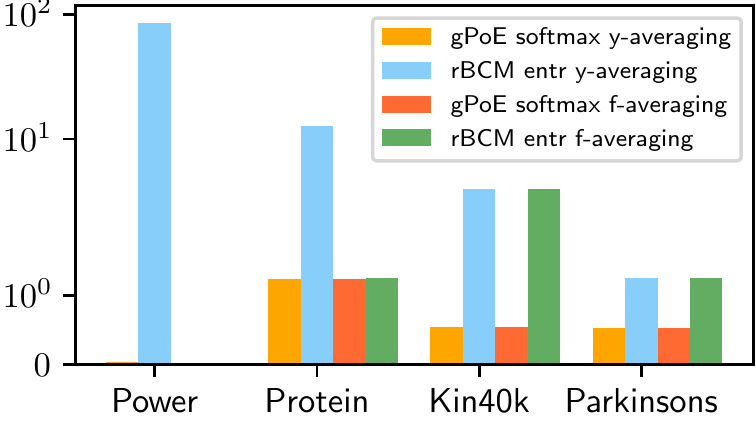}
  \caption{NLPD of the gPoE with our softmax proposal, and the rBCM with differential entropy under $f$- and $y$- averaging regimes (random partitioning). The latter model leads to weak performance under $y$-averaging whilst our proposal is robust in both regimes.}
  \label{fig:nlpd_prob}
\end{figure}  

In Figure \ref{fig:nlpd_prob}, we show that with random partitioning, the rBCM with differential entropy has weak performance using $y$-averaging (in particular on power and protein), which is similar to results reported in all three papers, whilst its performance is good when using $f$-averaging. This explains the discrepancy between the results in \cite{pmlr-v37-deisenroth15} and in our paper ($f$-averaging) and the results reported by \citet{liu2018generalized,trapp2019structured,JMLR:v20:18-374} ($y$-averaging). The problem with $y$-averaging has to do with weight calibration. We see in Figure \ref{fig:nlpd_prob} that our calibrating approach leads to strong performance in both $y$- and $f$- averaging, thus healing PoEs in both settings. 

\begin{figure}

\centering
\includegraphics[width=0.49\hsize]{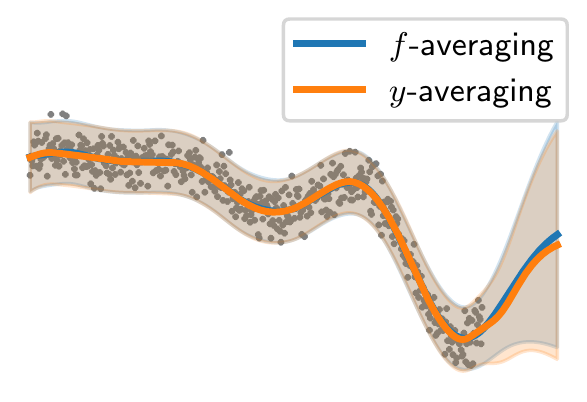}
\includegraphics[width=0.49\hsize]{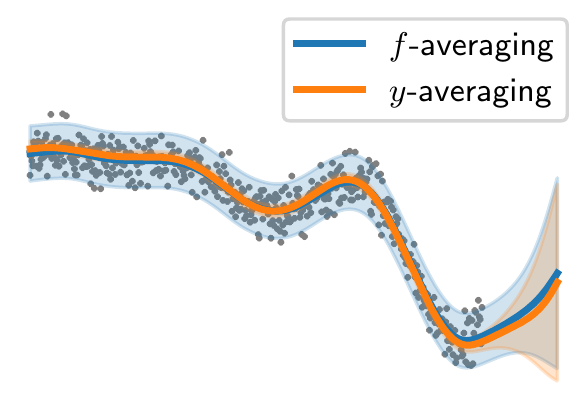}

  \caption{Predictions with BarGP softmax-var (Left) and rBCM diff-entropy (Right) in the $f$- and $y$- averaging regimes. The former works in both whilst the latter's variance shrinks under $y$-averaging}
  \label{fig:rob_bar_fy}
\end{figure}

We also provide an illustration of such results in Figure \ref{fig:rob_bar_fy}, using data from \cite{liu2018generalized}. We observe that when using differential entropic weighting, we recover the poor results reported \citet{liu2018generalized,JMLR:v20:18-374}, in which the variance shrinks significantly under $y$-averaging (whilst the model is sensible under $f$-averaging). By contrast, when using softmax-variance weighting, which calibrates the model, sensible results are obtained for both $f$- and $y$- averaging, further corroborating the quantitative results of Figure \ref{fig:nlpd_prob}. We argue that using normalised weights for the rBCM is essential. Otherwise, even in the case, where the number of experts is $M=1$, the rBCM does not have a predictive distribution equivalent to the one of a full GP; see Proposition \ref{prop:uneq} (Appendix), which gives an intuition for the erratic behaviours typically observed in the transitioning region for rBCM with unnormalised weights (e.g., Figure \ref{fig:rbcm}). We therefore advise practitioners to average in $f$-space, use normalised weighting approaches and appropriate weight calibration.

\section{Conclusion}
We identified significant shortcomings of previous approaches, notably the PoE, BCM, gPoE and rBCM, to scaling GP regression and classification via local-expert averaging. These models struggle in settings, where the number of strong experts is small, but the experts' weights are not sparse enough. Weight sparsity should thus be set to account for the overall strength of experts. To address these shortcomings, we control weight sparsity via the use of (normalised) softmax weights, along with a temperature to enforce this trade-off. Note that our approach can be combined with SVGPs \cite{pmlr-v38-hensman15} (as was done in Section \ref{sec:classification_exp}) but also with other methods, such as KISS-GP \cite{pmlr-v37-wilson15}. 
We provide strong empirical evidence that shortcomings of previous expert models can be addressed through this approach, which leads to substantial performance gains across benchmarks.
We further propose a novel scalable and distributable approach to averaging GP experts' predictions by means of Wasserstein barycenters, which can be used for regression and classification problems. When combined with our weighting proposal, it obtains state-of-the art performance across most datasets.




\subsection*{Acknowledgements}
We are grateful to James T. Wilson for constructive feedback on the paper. We would also like to thank Martin Trapp and Michael Zhang for providing code and helping in investigating characteristics of expert models.  SC is supported by the Engineering and Physical Sciences Research Council [grant number EP/S021566/1]. RM is supported by a Google PhD Fellowship in Machine Learning. TM is supported by the National Research Foundation of South Africa. 

\bibliographystyle{icml2020}
\bibliography{references}

\begin{thebibliography}{30}
\providecommand{\natexlab}[1]{#1}
\providecommand{\url}[1]{\texttt{#1}}
\expandafter\ifx\csname urlstyle\endcsname\relax
  \providecommand{\doi}[1]{doi: #1}\else
  \providecommand{\doi}{doi: \begingroup \urlstyle{rm}\Url}\fi

\bibitem[Bauer et~al.(2016)Bauer, van~der Wilk, and Rasmussen]{NIPS2016_6477}
Bauer, M., van~der Wilk, M., and Rasmussen, C.~E.
\newblock Understanding probabilistic sparse {G}aussian process approximations.
\newblock In \emph{NeurIPS}, 2016.

\bibitem[Bertone et~al.(2019)Bertone, Deisenroth, Kim, Liem, {Ruiz de Austri},
  and Welling]{Bertone2019}
Bertone, G., Deisenroth, M.~P., Kim, J.~S., Liem, S., {Ruiz de Austri}, R., and
  Welling, M.
\newblock Accelerating the {BSM} interpretation of {LHC} data with machine
  learning.
\newblock \emph{Physics of the Dark Universe}, 2019.

\bibitem[Bishop \& Svens{\'e}n(2003)Bishop and Svens{\'e}n]{bishop2012bayesian}
Bishop, C.~M. and Svens{\'e}n, M.
\newblock Bayesian hierarchical mixtures of experts.
\newblock In \emph{UAI}, 2003.

\bibitem[Bonneel \& Pfister(2013)Bonneel and Pfister]{bonneel2013sliced}
Bonneel, N. and Pfister, H.
\newblock Sliced {W}asserstein barycenter of multiple densities.
\newblock \emph{Harvard Technical Report}, 2013.

\bibitem[Cao \& Fleet(2014)Cao and Fleet]{Cao2014GeneralizedPO}
Cao, Y. and Fleet, D.~J.
\newblock Generalized product of experts for automatic and principled fusion of
  {G}aussian process predictions.
\newblock \emph{arXiv:1410.7827}, 2014.

\bibitem[Cao \& Fleet(2015)Cao and Fleet]{Cao2015TransductiveLO}
Cao, Y. and Fleet, D.~J.
\newblock Transductive log opinion pool of {G}aussian process experts.
\newblock \emph{arXiv:1511.07551}, 2015.

\bibitem[Csato \& Opper(2002)Csato and Opper]{soton259182}
Csato, L. and Opper, M.
\newblock Sparse online {G}aussian processes.
\newblock \emph{Neural Computation}, 2002.

\bibitem[Deisenroth \& Ng(2015)Deisenroth and Ng]{pmlr-v37-deisenroth15}
Deisenroth, M.~P. and Ng, J.~W.
\newblock Distributed {G}aussian processes.
\newblock In \emph{ICML}, 2015.

\bibitem[Gal et~al.(2014)Gal, van~der Wilk, and Rasmussen]{NIPS2014_5593}
Gal, Y., van~der Wilk, M., and Rasmussen, C.~E.
\newblock Distributed variational inference in sparse {G}aussian process
  regression and latent variable models.
\newblock In \emph{NeurIPS}, 2014.

\bibitem[Guo et~al.(2017)Guo, Pleiss, Sun, and Weinberger]{guo2017calibration}
Guo, C., Pleiss, G., Sun, Y., and Weinberger, K.~Q.
\newblock On calibration of modern neural networks.
\newblock In \emph{ICML}, 2017.

\bibitem[Hensman et~al.(2013)Hensman, Fusi, and
  Lawrence]{Hensman:2013:GPB:3023638.3023667}
Hensman, J., Fusi, N., and Lawrence, N.~D.
\newblock Gaussian processes for big data.
\newblock In \emph{UAI}, 2013.

\bibitem[Hensman et~al.(2015)Hensman, de~G.~Matthews, and
  Ghahramani]{pmlr-v38-hensman15}
Hensman, J., de~G.~Matthews, A.~G., and Ghahramani, Z.
\newblock Scalable variational {G}aussian process classification.
\newblock In \emph{AISTATS}, 2015.

\bibitem[Hinton(1999)]{hinton1999products}
Hinton, G.~E.
\newblock Products of experts.
\newblock In \emph{ICANN}, 1999.

\bibitem[Liu et~al.(2018)Liu, Cai, Wang, and Ong]{liu2018generalized}
Liu, H., Cai, J., Wang, Y., and Ong, Y.-S.
\newblock Generalized robust {B}ayesian committee machine for large-scale
  {G}aussian process regression.
\newblock \emph{arXiv:1806.00720}, 2018.

\bibitem[Platt(1999)]{platt1999probabilistic}
Platt, J.
\newblock Probabilistic outputs for support vector machines and comparisons to
  regularized likelihood methods.
\newblock \emph{Advances in Large Margin Classifiers}, 1999.

\bibitem[Qui\~{n}onero Candela \& Rasmussen(2005)Qui\~{n}onero Candela and
  Rasmussen]{10.5555/1046920.1194909}
Qui\~{n}onero Candela, J. and Rasmussen, C.~E.
\newblock A unifying view of sparse approximate {G}aussian process regression.
\newblock \emph{Journal of Machine Learning Research}, 2005.

\bibitem[Rasmussen \& Ghahramani(2001)Rasmussen and
  Ghahramani]{DBLP:conf/nips/RasmussenG01}
Rasmussen, C.~E. and Ghahramani, Z.
\newblock Infinite mixtures of {G}aussian process experts.
\newblock In \emph{NeurIPS}, 2001.

\bibitem[Rasmussen \& Williams(2006)Rasmussen and
  Williams]{Rasmussen:2005:GPM:1162254}
Rasmussen, C.~E. and Williams, C. K.~I.
\newblock \emph{Gaussian Processes for Machine Learning}.
\newblock The MIT Press, 2006.

\bibitem[Rulli{\`e}re et~al.(2018)Rulli{\`e}re, Durrande, Bachoc, and
  Chevalier]{rulliere:hal-01345959}
Rulli{\`e}re, D., Durrande, N., Bachoc, F., and Chevalier, C.
\newblock Nested {K}riging predictions for datasets with large number of
  observations.
\newblock \emph{{Statistics and Computing}}, 2018.

\bibitem[Snelson \& Ghahramani(2006)Snelson and Ghahramani]{NIPS2005_2857}
Snelson, E. and Ghahramani, Z.
\newblock Sparse {G}aussian processes using pseudo-inputs.
\newblock In \emph{NeurIPS}, 2006.

\bibitem[Srivastava et~al.(2015)Srivastava, Cevher, Dinh, and
  Dunson]{Srivastava2015}
Srivastava, S., Cevher, V., Dinh, Q., and Dunson, D.
\newblock {WASP: S}calable {B}ayes via barycenters of subset posteriors.
\newblock In \emph{AISTATS}, 2015.

\bibitem[Titsias(2009)]{Titsias09variationallearning}
Titsias, M.~K.
\newblock Variational learning of inducing variables in sparse {G}aussian
  processes.
\newblock In \emph{AISTATS}, 2009.

\bibitem[Trapp et~al.(2019)Trapp, Peharz, Pernkopf, and
  Rasmussen]{trapp2019structured}
Trapp, M., Peharz, R., Pernkopf, F., and Rasmussen, C.~E.
\newblock Deep structured mixtures of {G}aussian processes.
\newblock In \emph{AISTATS}, 2019.

\bibitem[Tresp(2000{\natexlab{a}})]{10.5555/3008751.3008843}
Tresp, V.
\newblock Mixtures of {G}aussian processes.
\newblock In \emph{NeurIPS}, 2000{\natexlab{a}}.

\bibitem[Tresp(2000{\natexlab{b}})]{Tresp2000}
Tresp, V.
\newblock A {B}ayesian committee machine.
\newblock \emph{Neural Computation}, 2000{\natexlab{b}}.

\bibitem[Villani(2008)]{Villani2008OptimalTO}
Villani, C.
\newblock \emph{Optimal Transport: Old and New}, volume 338.
\newblock Springer Science \& Business Media, 2008.

\bibitem[Wang et~al.(2019)Wang, Pleiss, Gardner, Weinberger, and
  Wilson]{Wang2019}
Wang, K.~A., Pleiss, G., Gardner, J.~R., Weinberger, K.~Q., and Wilson, A.~G.
\newblock Exact {G}aussian processes on a million data points.
\newblock In \emph{NeurIPS}, 2019.

\bibitem[Wilson \& Nickisch(2015)Wilson and Nickisch]{pmlr-v37-wilson15}
Wilson, A.~G. and Nickisch, H.
\newblock Kernel interpolation for scalable structured {G}aussian processes
  ({KISS-GP}).
\newblock In \emph{ICML}, 2015.

\bibitem[Zhang \& Williamson(2019)Zhang and Williamson]{JMLR:v20:18-374}
Zhang, M.~M. and Williamson, S.~A.
\newblock Embarrassingly parallel inference for {G}aussian processes.
\newblock \emph{Journal of Machine Learning Research}, 2019.

\bibitem[Álvarez Esteban et~al.(2016)Álvarez Esteban, del Barrio,
  Cuesta-Albertos, and Matrán]{pub.1023863441}
Álvarez Esteban, P.~C., del Barrio, E., Cuesta-Albertos, J.~A., and Matrán,
  C.
\newblock A fixed-point approach to barycenters in {W}asserstein space.
\newblock \emph{Journal of Mathematical Analysis and Applications}, 2016.

\end{thebibliography}



\appendix
\onecolumn
\section{Properties of Product-of-Experts}
\label{sec:apdxtheory}
We now discuss the properties of PoEs, and provide formal proofs for the theoretical statements made throughout the paper. We first prove Proposition \ref{prop:equivpoe}, which states that when weights are normalized, gPoE and rBCM are equivalent. We then prove Proposition \ref{prop:equivtemp}, which states that in the infinite temperature limit, under our proposed softmax-variance weighting, gPoE, rBCM and the barycenter of GPs are equivalent. 

\textit{Proof of Proposition \ref{prop:equivpoe}}: 

If weights are normalised, i.e. $\sum_j \beta_j(\v{x}_*) = 1 $ for all $\v{x}_*$ then it holds  
\[
\sigma^{-2}_{rbcm}(\v{x}_*)=\sum_{j=1}^J \beta_{j}(\v{x}_*)\sigma_j^{-2}(\v{x}_*) = \sigma^{-2}_{gpoe}(\v{x}_*),
\]
and the aggregated means are thus also equal $m_{gpoe}(\v{x}_*)=m_{rbcm}(\v{x}_*)$ for all $ \v{x}_*$. 

\textit{Proof of Proposition \ref{prop:equivtemp}}: 

We consider the limit $T\rightarrow \infty$, and we consider weighting the experts using $\beta_j(\v{x}_*)= \frac{e^{-T\sigma_j^2(\v{x}_*)}  }{\sum_k e^{-T\sigma_k^2(\v{x}_*)}} $ where $\sigma_j^2(\v{x}_*)$ is the predictive variance of expert $j$ at $\v{x}_*$. We set $\sigma^2_{\min}(\v{x}_*) = \text{min}\{\sigma_1^2(\v{x}_*),..., \sigma_J^2(\v{x}_*)\}$, and define $K$ as
\begin{align}
    K = | \{k \in \{1,...,J\}: \sigma_j^2(\v{x}_*) = \sigma^2_{\min}(\v{x}_*)\}|.
\end{align}

First, we consider the case  $\sigma_j^2(\v{x}_*) = \sigma^2_{\min}(\v{x}_*)$. Then
\begin{align}
     \lim\limits_{T\rightarrow \infty}\beta_j(\v{x}_*)&= \lim\limits_{T\rightarrow \infty}\frac{1}{\sum_k e^{T(\sigma_j^2(\v{x}_*)-\sigma_k^2(\v{x}_*))}}=\lim\limits_{T\rightarrow \infty} \frac{1}{Ke^{T(\sigma_j^2(\v{x}_*)-\sigma^2_{\min} (\v{x}_*)})}= \frac{1}{K}.
\end{align}

Second, we consider the case that $\sigma_j^2(\v{x}_*) > \sigma^2_{\min}(\v{x}_*)$. Then 
\begin{align}
    \lim\limits_{T\rightarrow \infty}\beta_j(\v{x}_*)&=\lim\limits_{T\rightarrow \infty}\frac{e^{-T\sigma_j^2(\v{x}_*)}  }{\sum_l e^{-T\sigma_l^2(\v{x}_*)}}
    = \lim\limits_{T\rightarrow \infty}\frac{1}{\sum_l e^{T(\sigma_j^2(\v{x}_*)-\sigma_l^2(\v{x}_*))}}=0,
\end{align}
since at least one expert has variance $\sigma_{\min}^2(\v{x}_*)<\sigma_j^2(\v{x}_*)$, and hence the term in the denominator goes to infinity.

Therefore, only experts that have minimum predictive variance $\sigma_j^2(\v{x}_*)=\sigma^2_{\min}(\v{x}_*)$ have (uniform) weight $\frac{1}{K}$ where $K$ is the number of experts having minimum variance.

From Proposition \ref{prop:equivpoe}, we know that under this weighting scheme, the gPoE and the rBCM are equivalent. We now show that the gPoE is equivalent to the barycenter of GPs, which will prove the result. 
Firstly, we have 
\begin{align}
    \sigma^{-2}_{gpoe}(\v{x}_*) &= \sum_j\beta_j(\v{x}_*)\sigma_j^{-2}(\v{x}_*) = \frac{1}{K}\sum_{j: \sigma_j^2 = \sigma^{2}_{\min}(\v{x}_*)} \sigma_{j}^{-2}(\v{x}_*) = \sigma^{-2}_{\min}(\v{x}_*),
\\
    \sigma^{2}_{bar}(\v{x}_*) &= \sum_j\beta_j(\v{x}_*)\sigma_j^{2}(\v{x}_*) = \frac{1}{K}\sum_{j: \sigma_j^2 = \sigma^2_{\min}(\v{x}_*)} \sigma_{j}^{2}(\v{x}_*) = \sigma^{2}_{\min}(\v{x}_*).
\end{align}
Thus, for $T\to\infty$ and with normalised weights, the gPoE, BarGP and the rBCM have the same predictive variance. We now show the equivalence of the predictive means:
\begin{align}
    \mu_{bar}(\v{x}_*)&=\sum\limits_{j=1}^J \beta_j(\v{x}_*)m_j(\v{x}_*)= \frac{1}{K}\sum_{j: \sigma_j^2(\v{x}_*) = \sigma^{2}_{\min}(\v{x}_*)}m_j(\v{x}_*)
    \\
    \mu_{gpoe}(\v{x}_*)&=\sigma^{2}_{\min}(\v{x}_*)\sum\limits_{j: \sigma_j^2(\v{x}_*) = \sigma^{2}_{\min}(\v{x}_*)} \frac{1}{K} \sigma_{j}^{-2}(\v{x}_*)m_j(\v{x}_*) \\&= \sigma^{2}_{\min} \sigma^{-2}_{\min}(\v{x}_*)\frac{1}{K}\sum_{j: \sigma_j^2(\v{x}_*) = \sigma^{2}_{\min}(\v{x}_*)}m_j(\v{x}_*)=\frac{1}{K}\sum_{j: \sigma_j^2(\v{x}_*) = \sigma^{2}_{\min}(\v{x}_*)}m_j(\v{x}_*),
\end{align}
which proves the result. \qed
\bigskip

Finally, we provide a results that illustrates a failure of rBCM with unnormalised weights, and provides intuition for the erratic behaviours of rBCM in the transitioning region.
\begin{proposition}
\label{prop:uneq}
If $J=1$, $0<\beta_1(\v{x}_*)\neq 1$, and $\sigma_{**}^2>\sigma_1^2(\v{x}_*)>0$ the rBCM is not equivalent to the full GP with identical hyperparameters.
\end{proposition}

\textit{Proof of Proposition \ref{prop:uneq}}: 
If $J=1$, the predictive precision of the rBCM is of the form
\begin{align}
    \sigma_{rbcm}^{-2}(\v{x}_*)=\beta_1(\v{x}_*)\sigma_{1}^{-2}(\v{x}_*)+(1-\beta_1(\v{x}_*))\sigma_{**}^{-2} \neq \sigma_{1}^{-2}(\v{x}_*).
\end{align}
Therefore, the mean is not identical either as the variance terms do not cancel.
\qed

\section{Extra Experimental Details}
All experts share the same hyperparameters, which are trained jointly by maximising the marginal likelihood for full GPs (regression) or the ELBO for SVGP experts. The classification SVGP experts use a multiclass likelihood with a robustmax link function, and RBF kernels with ARD. We use L-BFGS-B with a maximum of $100$ iterations for full GP, and ADAM for SVGP experts. 

We compare against SVGP \cite{Hensman:2013:GPB:3023638.3023667} with $500$ inducing points. 
For classification, we use random partitioning such that each expert is allocated to clusters with 500 training datapoints.

\section{Additional Experimental Results}

\subsection{Random Data Partitioning Results}

Table \ref{Tab:random_part} shows the regression results when using random data partitioning.

\begin{table*}[htb]
\resizebox{\linewidth}{!}{
\begin{tabular}{rrrrrrrrrr}
\toprule
	dataset & $N$ & $D$ &gPoE\_Unif	&	gPoE/rBCM\_var &		rBCM\_entr & BAR\_var &		linear		&$\text{SVGP}_{500}$	\\
	\midrule
Airfoil &1503&8&	0.820	(\blue{0.540 })	&	0.767	(\blue{0.512})	&	0.795	(\blue{0.530})	&	0.774	(\blue{0.515})	&	1.096	(\blue{0.721})	&	0.409	(\blue{0.353})	 \\
Concrete &1030&5&	0.614	(\blue{0.453})	&	0.558	(\blue{0.426})	&	0.630	(\blue{0.447})	&	0.562	(\blue{0.428})	&	0.953	(\blue{0.626})	&	0.289	(\blue{0.338})	\\
Kin40k &40000&8 &	1.079	(\blue{0.717})	&	0.517	(\blue{0.408})	&	3.947	(\blue{0.470})	&	0.513	(\blue{0.405})	&	1.419	(\blue{1.000})	&	0.124	(\blue{0.263})	\\
Parkinsons	&5878&20 & 1.030	(\blue{0.669})	&	0.521	(\blue{0.415})	&	1.245	(\blue{0.481})	&	0.520	(\blue{0.415})	&	1.282	(\blue{0.871})	&	0.554	(\blue{0.412})	\\
Power &9568&4	&0.032	(\blue{0.249})	&	0.030	(\blue{0.248})	&	0.028	(\blue{0.249})	&	0.032	(\blue{0.249})	&	0.098	(\blue{0.267})	&	-0.044	(\blue{0.231})	 \\
Protein &45730&9&	1.270	(\blue{0.857})	&	1.251	(\blue{0.841})	&	1.252	(\blue{0.846})	&	1.253	(\blue{0.843})	&	1.254	(\blue{0.848})	&	1.083	(\blue{0.715})	 \\
average &&&	0.808	(\blue{0.581})	&	0.607	(\blue{0.475})	&	1.316	(\blue{0.504})	&	0.609	(\blue{0.476})	&	1.017	(\blue{0.722})	&	0.402	(\blue{0.385})	\\
\bottomrule
\end{tabular}
}
\caption{Average NLPD (\blue{RMSE}) values across small (1K+ points) and large-scale (10k+ points) benchmarks when using random data partitioning.}
\label{Tab:random_part}
\end{table*}

\subsection{Robustness to the Temperature Parameter}

The performance of the expert models tends to vary depending on the  temperature parameter $T$.Tables \ref{Tab:var_bar_roburst}--\ref{Tab:var_rbcm_roburst} show similar performance metrics of the variance weighted expert models. A key observation is that the performance of the variance-weighted barGP and gPoE experts stabilise at temperatures above 10. This is desirable behaviour in terms of weight allocation across experts. 

\begin{table*}[htb]
\resizebox{\linewidth}{!}{
\begin{tabular}{rrrrrrrrr}
\toprule
	dataset &$\text{BAR\_var}_{T=1}$  & $\text{BAR\_var}_{T=10}$ & $\text{BAR\_var}_{T=25}$ & $\text{BAR\_var}_{T=50}$&$\text{BAR\_var}_{T=75}$ & $\text{BAR\_var}_{T=100}$ & $\text{BAR\_var}_{T=150}$ \\
	\midrule
Airfoil &	1.219	(\blue{0.729})	&	0.411	(\blue{0.350})	&	0.411	(\blue{0.349})	&	0.411	(\blue{0.350})	&	0.411	(\blue{0.350})	&	0.411	(\blue{0.350})	&	0.411	(\blue{0.350})	 \\
Concrete &	1.019	(\blue{0.488})	&	0.293	(\blue{0.344})	&	0.289	(\blue{0.343})	&	0.289	(\blue{0.342})	&	0.288	(\blue{0.342})	&	0.288	(\blue{0.342})	&	0.288	(\blue{0.342})	 \\
Kin40k &	1.217	(\blue{0.807})	&	0.106	(\blue{0.201})& -0.291	(\blue{0.158})	&	-0.365	(\blue{0.170})	&	-0.354	(\blue{0.178})	&	-0.339	(\blue{0.183})	&	-0.319	(\blue{0.190})	 \\
Parkinsons	& 1.342	(\blue{0.920})	&	0.208	(\blue{0.353})	&	0.093	(\blue{0.333})	&	0.093	(\blue{0.335})	&	0.097	(\blue{0.336	})	&	0.100	(\blue{0.337})	&	0.103	(\blue{0.339})	\\
Power &	1.017	(\blue{0.690})	&	0.239	(\blue{0.291})	&	0.025	(\blue{0.243})	&	-0.047	(\blue{0.229})	&	-0.068	(\blue{0.225})	&	-0.076	(\blue{0.224})	&	-0.082	(\blue{0.222})	 \\
Protein &	1.410	(\blue{0.992})	&	0.911	(\blue{0.676})	&	0.781	(\blue{0.602})	&	0.776	(\blue{0.587})	&	0.775	(\blue{0.583})	&		0.775	(\blue{0.582})	&	0.775	(\blue{0.580})	\\
average	& 1.204	(\blue{0.771})	&	0.361	(\blue{0.369})	&	0.218	(\blue{0.338})	&	0.193	(\blue{0.335})	&	0.192	(\blue{0.336})	&		0.193	(\blue{0.336})	&	0.196	(\blue{0.337})	\\
\bottomrule
\end{tabular}
}
\caption{Average NLPD (\blue{RMSE}) values across  benchmark datasets when using the posterior predictive variance  expert weights across various settings of the temperature parameter for the Barycenter prediction aggregation method.}
\label{Tab:var_bar_roburst}
\end{table*}

\begin{table*}[htb]
\resizebox{\linewidth}{!}{
\begin{tabular}{rrrrrrrrrr}
\toprule
	dataset &$\text{gPoE\_var}_{T=1}$  & $\text{gPoE\_var}_{T=10}$ & $\text{gPoE\_var}_{T=25}$ & $\text{gPoE\_var}_{T=50}$ &$\text{gPoE\_var}_{T=75}$ & $\text{gPoE\_var}_{T=100}$ & $\text{gPoE\_var}_{T=150}$\\
	\midrule
Airfoil	 &0.510	(\blue{0.395})	&	0.412	(\blue{0.350})	&0.411	(\blue{0.349})	&	0.411	(\blue{0.350})	&	0.411	(\blue{0.350})	&		0.411	(\blue{0.350})	&	0.411(\blue{0.350})	\\
Concrete &	0.309	(\blue{0.346})	&	0.290(\blue{	0.343})	&0.289	(\blue{0.342})	&	0.288	(\blue{0.342})	&	0.288	(\blue{0.342})		&	0.288	(\blue{0.342})	&	0.288	(\blue{0.342})	\\
Kin40k &	0.882	(\blue{0.550})	&	-0.240	(\blue{0.161})	&	-0.364	(\blue{0.164})	&	-0.359	(\blue{0.176})	&	-0.342	(\blue{0.182})		&	-0.329	(\blue{0.186})	&	-0.313	(\blue{0.191})	\\
Parkinsons &	0.855	(\blue{0.599})	&	0.094	(\blue{0.338})	&	0.088	(\blue{0.334})	&	0.094	(\blue{0.335})	&	0.098	(\blue{0.337})	&		0.101	(\blue{0.338})	&	0.104	(\blue{0.339})	\\
Power &	0.192	(\blue{0.279})	&	-0.052	(\blue{0.226})	&-0.076	(\blue{0.223})	&	-0.082	(\blue{0.222})	&	-0.083	(\blue{0.222})		&	-0.084	(\blue{0.222})	&	-0.084	(\blue{0.222})	\\
Protein &	1.344	(\blue{0.932})	&	0.821	(\blue{0.645})	&0.782	(\blue{0.601})	&	0.776	(\blue{0.587})	&	0.775	(\blue{0.583})	&		0.775	(\blue{0.582})	&	0.775	(\blue{0.580})	\\
average &	0.682	(\blue{0.517})	&	0.221	(\blue{0.344})	&0.188	(\blue{0.336})	&	0.188	(\blue{0.335})	&	0.191	(\blue{0.336})	&		0.194	(\blue{0.336})	&	0.197	(\blue{0.337})	\\
\bottomrule
\end{tabular}
}
\caption{Average NLPD (\blue{RMSE}) values across  benchmark datasets when using the posterior predictive variance  expert weights across various settings of the temperature parameter for the gPoE prediction aggregation method.}
\label{Tab:var_gpoe_roburst}
\end{table*}

\begin{table*}[htb]
\resizebox{\linewidth}{!}{
\begin{tabular}{rrrrrrrrrr}
\toprule		
dataset   & $\text{rBCM\_var}_{T=1}$ & $\text{rBCM\_var}_{T=10}$ & $\text{rBCM\_var}_{T=25}$&$\text{rBCM\_var}_{T=50}$ &$\text{rBCM\_var}_{T=75}$& $\text{rBCM\_var}_{T=100}$ & $\text{rBCM\_var}_{T=150}$ \\
\midrule
Airfoil &	0.412	(\blue{0.354})	&	0.497	(\blue{0.451})	&	0.583	(\blue{0.544})	&	0.674	(\blue{0.614})	&	0.754	(\blue{0.663})		&	0.833	(\blue{0.702})	&	0.985 (\blue{	0.768})	 \\
Concrete &	0.287	(\blue{0.344})	&	0.352	(\blue{0.405})	&0.431	(\blue{0.470})	&	0.547	(\blue{0.522})	&	0.665	(\blue{0.559})		&	0.775	(\blue{0.602})	&	0.947	(\blue{0.675})	 \\
Kin40k &	2.397	(\blue{0.290})	&	-0.539	(\blue{0.153})	&-0.271	(\blue{0.207})	&	0.124	(\blue{0.372})	&	0.423	(\blue{0.521})		&	0.640	(\blue{0.629})	&	0.916	(\blue{0.761})	 \\
Parkinsons	 & 0.302	(\blue{0.350})	&	0.209	(\blue{0.418})	&0.410	(\blue{0.528})	&	0.653	(\blue{0.651})	&	0.818	(\blue{0.731})		&	0.936	(\blue{0.785})	&	1.093	(\blue{0.856})	 \\
Power &	-0.064	(\blue{0.226})	&	-0.076	(\blue{0.223})	&	-0.080	(\blue{0.222})	&	-0.079		(\blue{0.222})	&	-0.074	(\blue{0.224})	&		-0.067	(\blue{0.227})	&	-0.049	(\blue{0.239})	 \\
Protein &	0.784	(\blue{0.564})	&	0.890	(\blue{0.696})	&	1.001	(\blue{0.769})	&	1.106	(\blue{0.828})	&	1.168	(\blue{0.860})		&	1.209	(\blue{0.881})	&	1.263	(\blue{0.909})	\\
average	 & 0.686	(\blue{0.355})	&	0.222	(\blue{0.391})	&0.346	(\blue{0.457})	&	0.504	(\blue{0.535})	&	0.626	(\blue{0.593})	&		0.721	(\blue{0.638})	&	0.859    (\blue{	0.701})	\\
\bottomrule
\end{tabular}
}
\caption{Average NLPD (\blue{RMSE}) values across  benchmark datasets when using the posterior predictive variance  expert weights across various settings of the temperature parameter for the rBCM prediction aggregation method with unnormalised weights.
}
\label{Tab:var_rbcm_roburst}
\end{table*}

\subsection{Sensitivity to the Number of Points per Expert}

Tables \ref{Tab:var_points_per_expert} shows the performances of the various expert models across different settings of the the number of initial points assigned per expert under posterior predictive variance weights.
The results show that the overall performance tends to improve as the number of points per expert increases while relative performance between models remains constant. 
\begin{table*}[htb]
\resizebox{\linewidth}{!}{
\begin{tabular}{rrrrrrrrrrr}
\toprule
Dataset & BAR\_var\_100p&BAR\_var\_200p&BAR\_var\_500p&gPoE\_var\_100p&gPoE\_var\_200&gPoE\_var\_500p&rBCM\_var\_100p&rBCM\_var\_200p&rBCM\_var\_500p\\
\midrule
Airfoil	& 0.411	(\blue{0.350})	&	0.401	(\blue{0.344})	&	0.363	(\blue{0.332})	&	0.411	(\blue{0.350})	&	0.401	(\blue{0.344})	&	0.363	(\blue{0.332})	&	0.674	(\blue{0.614})	&	0.641	(\blue{0.598})	&	0.536	(\blue{0.515})	 \\
Concrete &	0.289	(\blue{0.342})	&	0.277	(\blue{0.345})	&	0.261	(\blue{0.330})	&	0.288	(\blue{0.342})	&	0.277	(\blue{0.345})	&	0.261	(\blue{0.330})	&	0.547	(\blue{0.522})	&	0.503	(\blue{0.500})	&	0.381	(\blue{0.405})	 \\
Kin40k	& -0.365	(\blue{0.170})	&	-0.545	(\blue{0.140})	&	-0.720	(\blue{0.117})	&	-0.359	(\blue{0.176})	&	-0.557	(\blue{0.145})	&	-0.765	(\blue{0.120})	&	0.124	(\blue{0.372})	&	-0.322	(\blue{0.233})	&	-0.679	(\blue{0.164})	 \\
Parkinsons &	0.093	(\blue{0.335})	&	0.101	(\blue{0.331})	&	0.078	(\blue{0.333})	&	0.094	(\blue{0.335})	&	0.103	(\blue{0.332})	&	0.079	(\blue{0.333})	&	0.653	(\blue{0.651})	&	0.673	(\blue{0.657})	&	0.686	(\blue{0.679})	\\
Power &	-0.047	(\blue{0.229})	&	-0.039	(\blue{0.231})	&	-0.044	(\blue{0.231})	&	-0.082	(\blue{0.222})	&	-0.072	(\blue{0.225})	&	-0.066	(\blue{0.226})	&	-0.079	(\blue{0.222})	&	-0.070	(\blue{0.225})	&	-0.064	(\blue{0.226})	 \\
Protein &	0.776	(\blue{0.587})	&	0.805	(\blue{0.705})	&	0.814	(\blue{0.673})	&	0.776	(\blue{0.587})	&	0.805	(\blue{0.704})	&	0.814	(\blue{0.673})	&	1.106	(\blue{0.828})	&	1.085	(\blue{0.891})	&1.113	(\blue{0.888})	 \\
average &	0.193	(\blue{0.335})	&	0.167	(\blue{0.349})	&	0.125	(\blue{0.336})	&	0.188	(\blue{0.335})	&	0.159	(\blue{0.349})	&	0.114	(\blue{0.336})	&	0.504	(\blue{0.535})	&	0.418	(\blue{0.517})	&	0.329	(\blue{0.480})	 \\
\bottomrule
\end{tabular}
}
\caption{Average NLPD (\blue{RMSE}) values across  benchmark datasets when using the posterior predictive variance  expert weights across different settings for the number of initial points per expert.
}
\label{Tab:var_points_per_expert}
\end{table*}

  \begin{table*}[ht!]
  \centering
\scalebox{1}{
\begin{tabular}{llllll}
\toprule
    Concrete  &    Airfoil   &    Power &    Kin40k  &        Protein 
 \\
 \midrule
 0.352 (\blue{0.360})& 0.506 (\blue{0.396}) & -0.007 (\blue{0.238}) & -0.320 (\blue{0.153})& 0.838 (\blue{0.741})
   \\
\bottomrule
\end{tabular}
}
\caption{NLPDs/\blue{RMSEs} of the grBCM with $y$-averaging.}
\label{tab:yavergrbcm}
\end{table*}

\begin{figure*}[t]
\centering
  \subfigure[soft-var, 100 pts/exp]{
  \includegraphics[width=0.23\hsize]{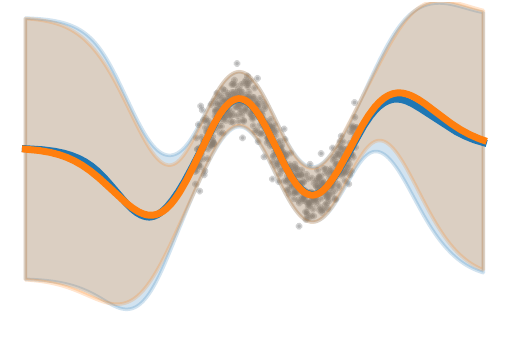}
  \label{fig:unif200_bar}
  }
  \subfigure[soft-var, 20 pts/exp]{
  \includegraphics[width=0.23\hsize]{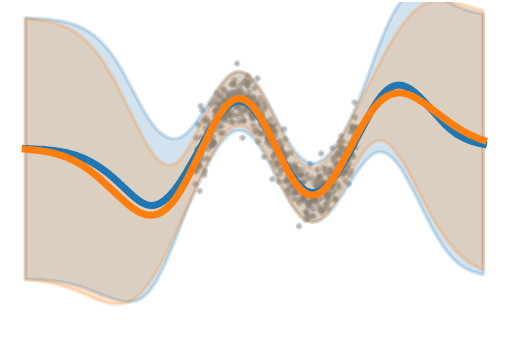}
  \label{fig:unif10_bar}
  }
 \subfigure[soft-var, 4 pts/exp]{
  \includegraphics[width=0.23\hsize]{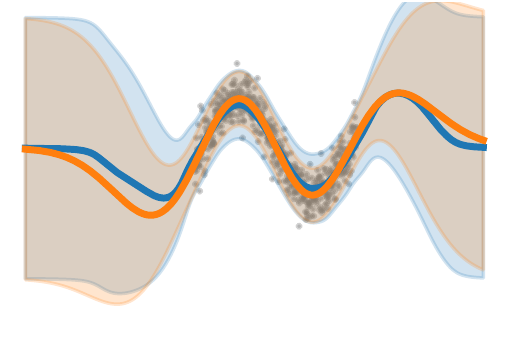}
  \label{fig:unif4_bar}
  }
  \subfigure[soft-var, 2 pts/exp]{
  \includegraphics[width=0.23\hsize]{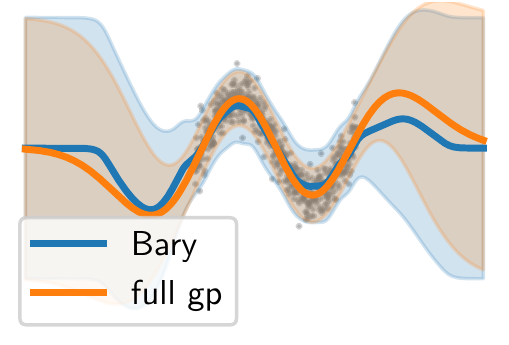}
  \label{fig:unif2_bar}
  
  }
\caption{Full GP baseline (orange) and barycenter of GPs model (blue) trained on synthetic data with a decreasing number of points per experts (Left to Right), using softmax-variance weighting.
}
\label{fig:robustness_bar}
\end{figure*}

\end{document}